\documentclass[conference]{IEEEtran}
\IEEEoverridecommandlockouts
\usepackage{cite}
\usepackage{booktabs}
\usepackage{amsmath,amssymb,amsfonts}
\usepackage{algorithmic}
\usepackage{graphicx}
\usepackage{textcomp}
\def\BibTeX{{\rm B\kern-.05em{\sc i\kern-.025em b}\kern-.08em
    T\kern-.1667em\lower.7ex\hbox{E}\kern-.125emX}}

\usepackage{xspace}
\usepackage[textsize=scriptsize]{todonotes}
\usepackage{tabularx}
\usepackage{myexample}
\usepackage{threeparttable}
\usepackage{url}
\usepackage{pifont}
\newcommand{\cmark}{\ding{51}}%
\newcommand{\xmark}{\ding{55}}%

\newcommand{\radtext}{RadText\xspace}

\begin{document}


\title{Radiology Text Analysis System (\radtext): Architecture and Evaluation\\
}

\author{
\IEEEauthorblockN{Song Wang\IEEEauthorrefmark{1},
Mingquan Lin\IEEEauthorrefmark{2},
Ying Ding\IEEEauthorrefmark{3},
George Shih\IEEEauthorrefmark{4},
Zhiyong Lu\IEEEauthorrefmark{5},
Yifan Peng\IEEEauthorrefmark{2}$^*$
}\\
\IEEEauthorblockA{\IEEEauthorrefmark{1}Cockrell School of Engineering, The University of Texas at Austin, Austin, USA}
\IEEEauthorblockA{\IEEEauthorrefmark{2}Department of Population Health Sciences,
Weill Cornell Medicine,
New York, USA}
\IEEEauthorblockA{\IEEEauthorrefmark{3}School of Information,
The University of Texas at Austin,
Austin, USA}
\IEEEauthorblockA{\IEEEauthorrefmark{4}Department of Radiology,
Weill Cornell Medicine,
New York, USA}
\IEEEauthorblockA{\IEEEauthorrefmark{5}National Center for Biotechnology Information (NCBI),
National Library of Medicine (NLM),\\
National Institutes of Health (NIH),
Bethesda, USA\\
$^*$Email: yip4002@med.cornell.edu}
}

\maketitle

\begin{abstract}
Analyzing radiology reports is a time-consuming and error-prone task, which raises the need for an efficient automated radiology report analysis system to alleviate the workloads of radiologists and encourage precise diagnosis. In this work, we present \radtext, an open-source radiology text analysis system developed by Python. \radtext offers an easy-to-use text analysis pipeline, including de-identification, section segmentation, sentence split and word tokenization, named entity recognition, parsing, and negation detection. \radtext features a flexible modular design, provides a hybrid text processing schema, and supports raw text processing and local processing, which enables better usability and improved data privacy. \radtext adopts BioC as the unified interface, and also standardizes the input / output into a structured representation compatible with Observational Medical Outcomes Partnership (OMOP) Common Data Model (CDM). This allows for a more systematic approach to observational research across multiple, disparate data sources. We evaluated \radtext on the MIMIC-CXR dataset, with five new disease labels we annotated for this work. \radtext demonstrates highly accurate classification performances, with an average precision of, a recall of 0.94, and an F-1 score of 0.92. 
We have made our code, documentation, examples, and the test set available at \url{https://github.com/bionlplab/radtext}.
\end{abstract}


\section{Introduction}

Radiology report analysis has long been a labor-some and error-prone process \cite{brady2016radiology-error}, which raises the need for accurate analysis tools to alleviate the workloads of radiologists and enhance accurate diagnosis. Though existing natural language processing (NLP) toolkits such as cTAKES \cite{savova2010ctakes}, scispaCy \cite{neumann-etal-2019-scispacy}, MedTagger \cite{liu2013medtagger}, and CLAMP \cite{soysal2017clamp} have been widely used in text mining of clinical narratives in electronic health record (EHR), none of these tools on the use of NLP in EHRs is specific to radiology domain. 

\begin{table}[]
\vspace{1em}
    \centering
    \begin{tabular}[width=\textwidth]{llcccc}
    \toprule
    System & Language & Raw-Text & Locally & Fully & Open\\
    &  & Processing & Process & Neural & Source\\
    \midrule
    MetaMap & Prolog/Java & \cmark & Hybrid & \xmark & \cmark\\
    cTakes  & Java & \cmark & \cmark &  \xmark & \cmark\\
    medspaCy & Python & \cmark & \cmark & \xmark & \cmark\\
    MedTagger & Java/C & \cmark & \cmark &  \xmark & \cmark\\
    CLAMP & Java & \cmark & \cmark & Hybrid & \xmark\\
    \midrule
    \radtext & Python & \cmark & \cmark & Hybrid & \cmark\\
    \bottomrule
    \end{tabular}
    \vspace{.5cm}
    \caption{Feature comparisons of \radtext against other widely used NLP toolkits. Fully Neural: full neural network pipeline.}
    \label{tab:comparison}
\end{table}

One recognized challenge is the requirement of proper radiology domain knowledge, without which the process of analyzing the structure of radiology text and interpreting the underlying meaning would be highly error-prone. For example, standardized terminology for each concept is important for NLP applications. Existing clinical NLP systems frequently use UMLS Methathesaurus as the medical lexicon\cite{bodenreider2004unified}. However, few support RadLex, which offers radiology-specific terms such as devices and imaging techniques \cite{langlotz2006radlex}. As a result, ambiguous terms (e.g., acronyms) can be interpreted differently. Another example is negation detection, which is also essential in radiology because diagnostic imagining is often used to rule out a condition. Systems in the clinical domain frequently implement this functionality by combining manually crafted rules with key terms based on the syntactic analysis\cite{chapman2013extending, chapman2011documentlevel}. While they usually achieve good results in the general clinical domain, most cannot be directly applied to radiology reports mostly because sentences in radiology reports are usually telegraphic, with missing subjects and verbs. In addition, sentences in the radiology reports also contain long, complicated noun phrases. These obstacles pose a challenge to existing parsers that are modeled over well-formed sentences \cite{fan2013syntactic}. Therefore, the performance of negation detection algorithms significantly drops\cite{peng2017negbio} in the case of radiology reports. In such cases, filling in the gaps requires additional rules to handle ill-formed sentences.

Another challenge is that every software intends to perform tasks on data in various formats. It thus remains challenging to seamlessly interchange data in and between different NLP tools. Such a bottleneck prevents combining these tools into a larger, more powerful, and more capable system in the clinical domain. 
To bridge this gap, the Observational Medical Outcomes Partnership (OMOP) Common Data Model (CDM) is proposed to harmonize disparate observational databases of EHR \cite{voss2015CDM}. The goal is to transform data contained within those databases into a common format (data model) and representation (terminologies, vocabularies, coding schemes) so that systematic analyses can be conducted in the common format. While OMOP CDM is an excellent schema to store structured data and provides a \texttt{NOTE\_NLP} table to store NLP final results, it does not support representing complex, messy data between different NLP modules, such as hierarchical note structure  (section, passage, sentence, token). Furthermore, it is almost impossible to store the parsing trees of each sentence in \texttt{NOTE\_NLP} table. However, such text-preprocessing information is frequently reused in NLP algorithms and should be interchangeable and reusable. In addition, OMOP CDM must be realized in a relational database, which most of the common NLP tools do not support. These limitations result in the main barrier to the reuse of tools and modules and the development of text mining pipelines customized for different workflows.
One alternative solution is the BioC format \cite{comeau2013bioc}, an XML-based simple format to share text data and annotations. Unlike OMOP CDM, BioC emphasizes simplicity, interoperability, broad use and reuse of data interchange. It is thus suitable to represent, store and exchange the NLP results, especially complex intermediate results, in a simple manner. However, as initially designed for sharing different annotations relevant for biomedical research, BioC cannot be directly used for clinical notes. To overcome this issue, we propose to extend the BioC format with the OMOP CDM  schema, called BioC-CDM, to store the results generated in the annotation process of clinical NLP that can be easily converted and imported into OMOP CDM.


In this work, we present \radtext, an open-source Python radiology text analysis system. Unlike previous methods, \radtext features a hybrid text analysis pipeline that utilizes high-performance third-party implementations, including machine learning-based methods and rule-based methods. As shown in Table \ref{tab:comparison}, compared to existing widely-used NLP toolkits, \radtext has the following advantages:

\begin{itemize}
    \item \textbf{Unified Interface}. \radtext uses BioC-CDM format as the unified interface throughout the system pipeline. BioC format simplifies data representation and data exchange and satisfies all the NLP task requirements in \radtext.
    \item \textbf{Compatible with OMOP CDM}. \radtext standardizes its outputs into a structured representation compatible with OMOP CDM. This allows for transforming data into a common representation and further enables a systematic analysis of disparate observational data sources.
    \item \textbf{Easy to Use}. \radtext provides a user-friendly  interface. \radtext sequentially runs de-identification, section segmentation, sentence split, word tokenization, named entity recognition, parsing, and negation detection. Modular choice of design greatly improves flexibility, which enables users to adjust any module according to their specific use case, and to re-run each module if needed. 
    \item \textbf{Raw Text Processing}. \radtext takes raw text as input, which means no text preprocessing (e.g., tokenization, annotation) is needed. This greatly enhances the usability and generalizability of \radtext.
    \item \textbf{Local Machine}. The entire system pipeline of \radtext is running locally on CPU machines. No data will be uploaded to remote servers, greatly preserving user data privacy. 
    \item \textbf{Open Source}. To facilitate and drive future clinical NLP research and applications, \radtext is fully open source. We make the source code, documentation, examples, and human-annotated test set publicly available.
\end{itemize}

\section{Related Work}
Various NLP toolkits have been introduced to the clinical NLP community\cite{pons2016naturala} and have been successfully applied to the information extraction task from clinical text. MetaMap \cite{aronson2010metamap} uses a knowledge-intensive approach based on symbolic, NLP, and computational-linguistic techniques to map the biomedical text into the Unified Medical Language System (UMLS) Metathesaurus \cite{bodenreider2004umls}. Apache Clinical Text Analysis and Knowledge Extraction System (cTAKES) focuses on extracting clinical information from electronic health record free text, including processing clinical notes, and identifying clinical named entities \cite{savova2010ctakes}. Different from MetaMap and Apache cTAKES, which utilize machine learning methods to map words to medical concepts, MedTagger for indexing is built upon a fast string matching algorithm leveraging lexical normalization \cite{liu2013medtagger}. It thus requires rules designing and expert knowledge engineering. Instead of conducting sole information extraction, medspaCy \cite{eyre2021medspacy} and Clinical Language Annotation, Modeling and Processing (CLAMP) \cite{soysal2017clamp} are designed to be modularized so that users can choose from various choices of modular components for their individual applications. medspaCy features performing clinical NLP and text processing tasks with the popular spaCy \cite{honnibal2020spacy} framework, which provides a robust architecture for building and sharing custom, high-performance NLP pipelines \cite{eyre2021medspacy}. CLAMP also highlights enabling users to quickly build customized NLP pipelines for their clinical NLP tasks. Distinguished from these previous works, \radtext aims to provide a high-performance clinical NLP toolkit in Python that focuses on radiology text analysis. \radtext hence adopts a hybrid radiology text processing pipeline, bringing together a number of third-party analysis tools in the radiology domain, with each tool implementing one or more components of \radtext's working pipeline.

\section{System Design and Architecture}

\subsection{BioC-CDM: BioC format compatible with OMOP CDM}\label{bioc}

\begin{table*}[ht]
    \centering
    \begin{threeparttable}[b]
    \begin{tabularx}{.99\textwidth}{lllX}
    \toprule
    OMOP CDM field & BioC field & BioC class & Description\\
    \midrule
note\_nlp\_id & id & annotation & A unique identifier for each term extracted from a note.\\
note\_id & doc & document & A foreign key to the Note table, uniquely identifying the note.\\
section\_concept\_id & section\_concept\_id & passage & A foreign key to the predefined Concept in the Standardized Vocabularies representing the section of the extracted term.\\
snippet & - & - & A small window of text surrounding the term. \\
offset &  offset & \begin{tabular}[t]{@{}l@{}}passage\\sentence\\ annotation\end{tabular} & Character offset of the extracted term in the input note. \\
lexical\_variant & text & annotation & Raw text extracted by the NLP tool. \\
note\_nlp\_concept\_id & lemma & annotation & A foreign key to a Concept table, representing the normalized concept of the extracted term.\\
note\_nlp\_source\_concept\_id & source\_concept\_id & annotation & A foreign key to a Concept table that refers to the code in the source vocabulary used by the NLP system.\\
nlp\_system & nlp\_system & collection & Name and version of the NLP system that extracted the term.\\
nlp\_date,nlp\_date\_time & date & collection & The date of the note processing. \\
    
   term\_exists & exists\tnote{1} & annotation & If the patient actually has or had the condition.\\
term\_temporal & temporal & annotation & If a condition is “present” or just in the “past”.\\
    term\_modifiers & modifiers & annotation & Describes compactly all the modifiers extracted by the NLP system.\\
    \bottomrule
    \end{tabularx}
    \begin{tablenotes}
    \item [1] currently called ``negation''
    \end{tablenotes}
    \caption{Mapping radiology notes to the OMOP CDM and BioC using \radtext.}
    \label{tab:bioc-cdm}
    \end{threeparttable}
\end{table*}

We propose BioC-CDM to store the results generated in the annotation process of clinical NLP in the BioC format that can be easily converted and imported into OMOP CDM. A BioC-format file is an XML document as the basis of data class representation and data exchange, which can satisfy the needs of \radtext's NLP tasks throughout the entire pipeline\cite{comeau2013bioc}. OMOP CDM harmonizes disparate coding systems to a standardized vocabulary with minimal information loss. As a result, adopting BioC-CDM as \radtext's unified interface and using it as a common format representing all modular components’ output eliminates the barrier of integration and greatly enhances \radtext's interoperability. 
Table \ref{tab:bioc-cdm} shows the current and our proposed mappings between OMOP CDM and BioC. Section \ref{api-usage-convert} shows how \radtext can be used to implement mutual conversion between BioC format and OMOP CDM.

\subsection{Pipeline}

\begin{figure}[h]
    \centering
    \includegraphics[width=0.35\textwidth]{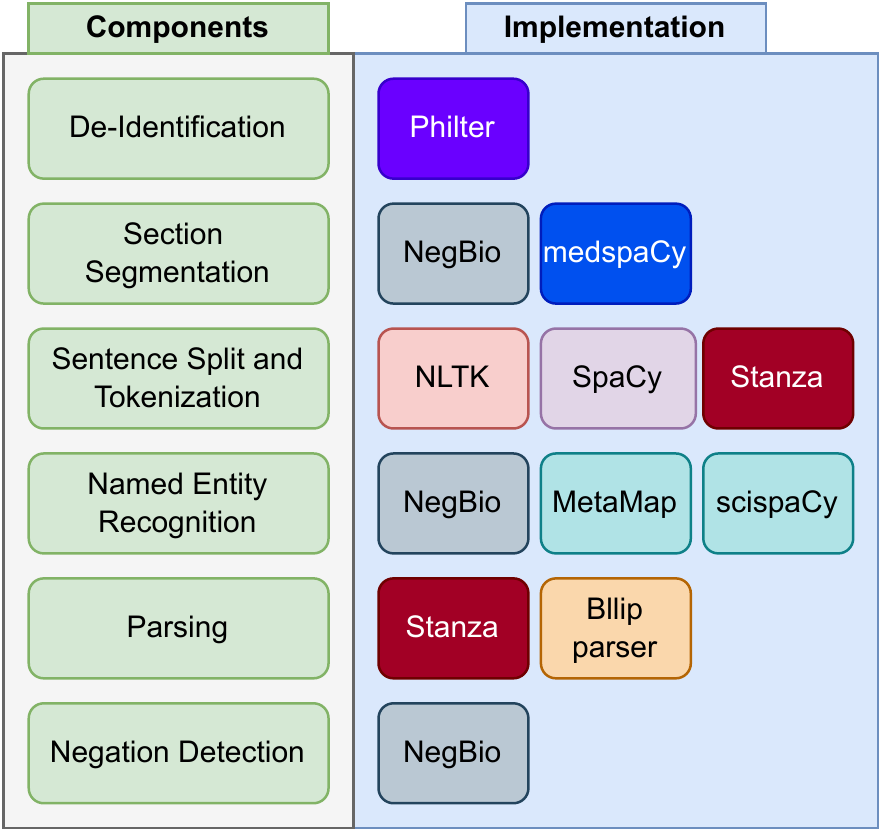}
    \caption{Overview of \radtext's NLP pipeline, main components, and implementations.}
    \label{fig:architecture}
\end{figure}

The implementation of \radtext is highly modular (Figure \ref{fig:architecture}). We highlight the details of each module in this section.

\subsubsection{De-Identification}

Radiology reports often contain protected health information (PHI), such as patient and provider names, addresses, and numbers\cite{Norgeot2020ProtectedHI}. Removal of PHI is important; otherwise, radiology reports remain largely unused for research. To address this issue, \radtext uses Philter \cite{Norgeot2020ProtectedHI} for de-identification. It uses both rule-based and statistical approaches to remove identifiers defined in the HIPAA Safe Harbor guidelines \cite{rightsocr2012guidance}. 

The following code snippet shows an example of \radtext's de-identification output. The mentions of patient's name, provider's name, and dates belong to PHI. They are replaced with a sequence of ``X"s respectively for de-identification purposes.

\begin{testexample}[Output in BioC]
\lstset{language=XML}
\begin{lstlisting}
<infon key="nlp_system">Philter</infon>
<document>
  <passage>
    <text>Patient's Name: XXXXXXXXXXXXXX
    Referred by: XXXXXXXXXXX XX
    Date Taken: XXXXXXXXXX
    Date of Report: XXXXXXXXXX
  
    Clinical statement: Shortness of breath, wheezing, and bilateral lower extremity edema.
    Technique: AP and lateral chest radiographs.
  
    Comparison: XXXXXXXXXXXXX...</text>
    <annotation id="A0">
      <infon key="source_concept">Date</infon>
      <infon key="source_concept_id">C1547350</infon>
      <location offset="70" length="10"/>
      <text>02/07/2016</text>
    </annotation>
    <annotation id="A1">
      <infon key="source_concept">Date</infon>
      <infon key="source_concept_id">C1547350</infon>
      <location offset="97" length="10"/>
      <text>02/07/2016</text>
    </annotation>
    <annotation id="A2">
      <infon key="source_concept">Date</infon>
      <infon key="source_concept_id">C1547350</infon>
      <location offset="263" length="13"/>
      <text>July 18, 2015</text>
    </annotation>
    <annotation id="A5">
      <infon key="source_concept">Person Name</infon>
      <infon key="source_concept_id">C1547383</infon>
      <location offset="16" length="14"/>
      <text>LATTE, MONICA</text>
    </annotation>
    <annotation id="A6">
      <infon key="source_concept">Person Name</infon>
      <infon key="source_concept_id">C1547383</infon>
      <location offset="43" length="11"/>
      <text>SAVEM, CARL</text>
    </annotation>
    <annotation id="A7">
      <infon key="source_concept">Degree/license/certificate</infon>
      <infon key="source_concept_id">C1547754</infon>
      <location offset="55" length="2"/>
      <text>MD</text>
    </annotation>
  </passage>
  ...
</document>
\end{lstlisting}
\end{testexample}

\subsubsection{Section Segmentation}

Although radiology reports are in the form of free text, they are often structured in terms of sections, such as INDICATION, FINDINGS, and IMPRESSION. Identifying section types and section boundaries can help various successive processing steps to use a subset of sections or assign specific weights to the content of different sections \cite{tepper2012statistical}. For example, effusion and edema were mentioned in the INDICATION section of the sample report below. But we should not identify them as positive because the radiologist ruled them out in the FINDINGS section. Therefore, a named entity recognition tool that does not differentiate between sections will likely make errors.


\begin{testexample}[An example of chest x-ray report]
\small
\begin{alltt}
{INDICATION}: Please evaluate for pneumonia, 
   {effusions, edema}
{FINDINGS}: The lungs are clear without 
   consolidation, {effusion or edema}...
{IMPRESSION}: No acute cardiopulmonary process.
\end{alltt}
\end{testexample}

In a preprocessing step, \radtext splits each report into sections and provides two options: NegBio or medspaCy. Both approaches rely on hand-coded heuristics for section segmentation (boundary detection) and achieve good performances. 




\begin{itemize}
\item \textbf{NegBio}. The heuristics in NegBio are based on conventions like the capitalization of headers and the presence of colon and blank lines between headers and text. The set of heuristics was collected from the NIH Chest X-ray dataset\cite{Wang2017ChestXRay8HC} and the MIMIC-CXR dataset\cite{johnson2019mimic}.

\item \textbf{medspaCy}. medspaCy includes an implementation of clinical section detection based on rule-based matching of the section titles with the default rules adapted from SecTag \cite{denny2008sectag} and expanded through practice. The default rules were collected from different resources such as the Logical Observation Identifiers Names and Codes (LOINC) headers \cite{mcdonald2003loinc} and Quick Medical Reference (QMR) Findings Hierarchy \cite{miller1989use} and were further revised based on the actual clinical notes from Vanderbilt EHR.
\end{itemize}

The following code snippet shows an example of the section segmentation output for the sample report above.

\begin{testexample}[Output in BioC]
\begin{lstlisting}[language=XML2]
<infon key="nlp_system">NegBio</infon>
<document>
  <passage>
    <infon key="section_concept">clinical information section </infon>
    <infon key="section_concept_id">RID13166</infon>
    <offset>0</offset>
    <text>INDICATION:</text>
  </passage>
  <passage>
    <offset>12</offset>
    <text>Please evaluate for ... edema</text>
  </passage>
  <passage>
    <infon key="section_concept">observations section</infon>
    <infon key="section_concept_id">RID28486</infon>
    <offset>60</offset>
    <text>FINDINGS:</text>
  </passage>
  <passage>
    <offset>70</offset>
    <text>The lungs are clear ... edema</text>
  </passage>
  ...
</document>
\end{lstlisting}
\end{testexample}

\subsubsection{Sentence Split and Word Tokenization}
\radtext tokenizes the input raw text and groups tokens into sentences as one part of preprocessing. \radtext offers three options to tokenize and split reports into sentences, including NLTK \cite{bird2009NLTK}, spaCy \cite{honnibal2020spacy}, and Stanza \cite{qi2020stanza}.

\begin{itemize}
\item \textbf{NLTK}. The sentence tokenizer in NLTK uses an unsupervised algorithm to build a model for abbreviation words, collocations, and words that start sentences. It then uses that model to find the sentence boundaries \cite{bird2009NLTK}.

\item \textbf{spaCy}. Sentence segmentation is part of spaCy's English pipeline. It uses a variant of the non-monotonic arc-eager transition system \cite{honnibal-johnson-2015-improved} with the addition of a ``break" transition for sentence segmentation \cite{honnibal2020spacy}. 

\item \textbf{Stanza}. Stanza combines tokenization and sentence segmentation from the raw text as one single module in its pipeline. Stanza models it  as a tagging task over character sequences, where the model predicts whether a given character is the end of a token, end of a sentence, or end of a multi-word token.
\end{itemize}

The following code snippet gives an example of \radtext's sentence split output. The input paragraph is split into three \texttt{Sentence} instances.

\begin{testexample}[Output in BioC]
\lstset{language=XML2}
\begin{lstlisting}
<infon key="nlp_system">NLTK</infon>
<document>
  <passage>
    <text>PA and lateral radiographs demonstrate clear lungs. Heart size is normal. There is no pneumothorax or pleural effusion.</text>
    <sentence>
      <offset>0</offset>
      <text>PA and lateral ... clear lungs.</text>
    </sentence>
    <sentence>
      <offset>52</offset>
      <text>Heart size is normal.</text>
    </sentence>
    <sentence>
      <offset>73</offset>
      <text>There is no ... pleural effusion.</text>
    </sentence>
  </passage>
  ...
</document>
\end{lstlisting}
\end{testexample}

\subsubsection{Named Entity Recognition}
Named entity recognition (NER) aims to determine and identify the words or phrases in text into predefined labels that describe the concepts of interest in a given domain \cite{nadeau2007ner}. To recognize the radiology-domain named entities (e.g., thoracic disorders) in each input sentence, \radtext offers two options, spaCy-enabled rule-based method and MetaMap.

\begin{itemize}
\item \textbf{Rule-based Regular Expression}. Rule-based NER methods use regular expressions that combine information from terminological resources and characteristics of the entities of interest manually constructed from report corpus. \radtext adopts spaCy's PhraseMatcher as part of this component. Rules defining concepts specify the text regular patterns to be matched and additional information about a concept, such as its unique id in the terminology.

\item \textbf{MetaMap}. UMLS is the most comprehensive standard terminology that is typically used as the basis for clinical concept extraction. Enabled by MetaMap, \radtext is able to detect all the concepts in UMLS and map them to Concept Unique Identifier (CUI). In general, MetaMap is much more comprehensive than vocabulary-based patterns. But at the same time, MetaMap could be noisy and less accurate. 

\end{itemize}

The following code snippet shows an example of \radtext's NER output, where ``Pneumonia" and ``Pneumothorax" are correctly recognized and their corresponding UMLS concept IDs are also identified.

\begin{testexample}[Output in BioC]
\lstset{language=XML}
\begin{lstlisting}
<infon key="nlp_system">MetaMap</infon>
<document>
  <passage>
    <text>There is no pneumonia or pneumothorax.</text>
    <annotation id="a1">
      <infon key="source_concept">Pneumonia</infon>
      <infon key="source_concept_id">RID5350</infon>
      <location offset="12" length="9"/>
      <text>pneumonia</text>
    </annotation>
    <annotation id="a2">
      <infon key="source_concept">Pneumothorax</infon>
      <infon key="source_concept_id">RID5352</infon>
      <location offset="24" length="12"/>
      <text>pneumothorax</text>
    </annotation> 
  </passage>
  ...
</document>
\end{lstlisting}
\end{testexample}

\subsubsection{Parsing}

\begin{figure}[h]
    \centering
    \includegraphics[width=0.5\textwidth]{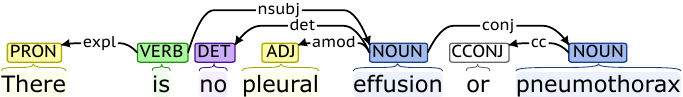}
    \caption{The obtained dependency graph of ``There is no pleural effusion or pneumothorax" using Stanza \cite{qi2020stanza}.}
    \label{fig:ud}
\end{figure}

\radtext utilizes the universal dependency graph (UDG) to describe the grammatical relationships within a sentence in a way that can be understood by non-linguists and effectively used by downstream processing tasks \cite{peng2017negbio}. UDG is a directed graph, which represents all universal dependency information within a sentence. The vertices in a UDG represent the information such as the word, lemma, and part-of-speech tag. The edges in a UDG represent the typed dependencies from the governor to its dependent and are labeled with the corresponding dependency type. UDG effectively represents the syntactic head of each word in a sentence and the dependency relations between words. Figure \ref{fig:ud} shows a UDG example of the sentence ``There is no pleural effusion or pneumothorax" generated by Stanza \cite{qi2020stanza}. In this example, ``pleural" is the adjectival modifier of ``effusion" and ``effusion" and ``pneumothorax" are coordinated findings.

To obtain the UDG of a sentence, \radtext provides two options, Stanza or Bllip Parser with the Stanford dependencies converter \cite{charniak-johnson-2005-coarse}.

\begin{itemize}
\item \textbf{Stanza}. Stanza's dependency parsing module builds a tree structure of words from the input sentence, representing the syntactic dependency relations between words. After \emph{tokenization}, \emph{multi-word token (MWT) expansion}, \emph{part-of-speech (POS) and morphological features tagging}, and \emph{lemmatization}, each sentence would have been directly parsed into the universal dependencies structure \cite{qi2020stanza}. 

\item \textbf{Bllip Parser with Stanford dependencies converter}. \radtext first parses each sentence to obtain the parse tree using the Bllip parser, which was trained with the biomedical model \cite{charniak-johnson-2005-coarse, Charniak2010AnyDP}. It then applies the Stanford dependencies converter on the resulting parse tree with the \emph{CCProcessed} and \emph{Universal} option \cite{de-marneffe-etal-2014-universal, marneffe2008dependency} to derive the universal dependencies.
\end{itemize}

The following code snippet shows an example of \radtext's parsing result. In the sample sentence, ``effusion" and ``pneumothorax" are respectively assigned with node id of ``T31" and ``T33". Derived from the universal dependency result, there is a conjunction relation between ``T31" and ``T33".

\begin{testexample}[Output in BioC]
\lstset{language=XML}
\begin{lstlisting}
<infon key="nlp_system">Bllip Parser</infon>
<document>
  <passage>
    <sentence>
      <infon key="parse tree">(S1 (S (S (NP (EX There)) (VP (VBZ is) (ADVP (RB no)) (NP  (NP) (JJ pleural) (NN effusion)) (CC or) (NP (NN pneumothorax))))) (. .)))</infon>
      <text>There is no pleural effusion or pneumothroax.</text>
      ...
      <annotation id="T31">
        <text>effusion</text>
      </annotation>
      <annotation id="T33">
        <text>pneumothorax</text>
      </annotation>
      ...
      <relation id="R33">
        <infon key="dependency">conj</infon>
        <node refid="T33" role="dependant"/>
        <node refid="T31" role="governor"/>
      </relation>
      ...
    </sentence>
    ...
  </passage>
  ...
</document>
\end{lstlisting}
\end{testexample}

\subsubsection{Negation Detection}
Negative and uncertain medical findings are frequent in radiology reports \cite{Chapman2001EvaluationON}. Since they may indicate the absence of findings mentioned within the radiology report, identifying them is as important as identifying positive findings. For negation and uncertainty detection, \radtext employs NegBio \cite{peng2017negbio, Wang2017ChestXRay8HC}, which utilizes universal dependencies for pattern definition and subgraph matching for graph traversal search so that the scope for negation/uncertainty is not limited to the fixed word distance \cite{de-marneffe-etal-2014-universal}.

The following code snippet shows an example of \radtext's negation detection output. In this sample sentence, ``pneumothorax" is identified as negative according to NegBio's internal negation rule of ID ``nn180".

\begin{testexample}[Output in BioC]
\lstset{language=XML}
\begin{lstlisting}
<infon key="nlp_system">NegBio</infon>
<document>
  <passage>
    <text>There is no pneumonia or pneumothorax.</text>
    ...
    <annotation id="a2">
      <infon key="source_concept">Pneumothorax</infon>
      <infon key="source_concept_id">RID5352</infon>
      <infon key="exists">False</infon>
      <infon key="negation">True</infon>
      <infon key="negbio_pattern_id">nn180</infon>
      <infon key="negbio_pattern_str">{}=f &gt;{} {lemma:/no/}=k0</infon>
      <location offset="24" length="12"/>
      <text>pneumothorax</text>
    </annotation>
  </passage>
  ...
</document>
\end{lstlisting}
\end{testexample}

\section{System Usage}
\radtext is designed to have a user-friendly interface and allow quick out-of-the-box usage for radiology text analysis. To achieve this, \radtext provides automated pipeline usage and step-by-step modular choice of design. 
Therefore, Users can run \radtext directly through the command line interface or import \radtext as a Python library to use any functionality through \radtext's API.

\subsection{Installation}

The latest \radtext releases are available on PyPI \footnote{\url{https://pypi.org/project/radtext/}}.
Using pip, \radtext releases can be downloaded as source packages and binary wheels. It is also generally recommended installing \radtext packages in a virtual environment to avoid modifying system state:

\begin{testexample}[Installation instructions]
\begin{lstlisting}[language=bash]
$ python -m venv venv
$ source venv/bin/activate
$ pip install -U radtext
$ python -m spacy download en_core_web_sm
$ radtext-download --all
\end{lstlisting}
\end{testexample}

\subsection{Command Line Usage}

The following command runs \radtext's entire pipeline in the sequential order of de-identification, section segmentation, sentence split and word tokenization, NER, parsing, and negation detection. The default section title vocabulary for the section segmentation module and concept vocabulary for the NER module is designed to be configurable. All intermediate result files will be generated and saved for use and reuse. The automatic pipeline execution enables users to use \radtext as an out-of-the-box toolkit without the need and effort to figure out how each module of \radtext works.

\begin{testexample}[An example of command line usage]
\begin{lstlisting}[language=bash]
$ bash run_pipeline.sh
\end{lstlisting}
\end{testexample}

In addition to running \radtext's pipeline as a whole, users can also choose to run every single module of \radtext through easy-to-use command line commands (see Table \ref{tab:radtext-commands}). This enables users to re-run each single modular component to reproduce the result in case of any error, without the need of re-running \radtext's entire pipeline. All intermediate results are saved so that users can easily check the output of each module, which we believe will greatly facilitate error analysis and enhance \radtext's flexibility. The following code snippet shows a an example of \radtext's modular command line usage.

\begin{testexample}[An example of modular command line usage]
\begin{lstlisting}[language=bash]
$ [command] [options] -i INPUT -o OUTPUT
$ radtext-deid -i /path/to/input.xml -o /path/to/output.xml
\end{lstlisting}
\end{testexample}

\begin{table}[h]
    \centering
    \begin{tabular}{ll}
    \toprule
    Commands & Description \\
    \midrule
    radtext-download & Download all models needed.\\
    radtext-deid & De-identifies all the reports.\\
    radtext-secsplit & Segments sections. \\
    radtext-ssplit & Splits sentences and tokenizes words. \\
    radtext-ner & Recognizes named entities.\\
    radtext-parse & Parses the sentences to obtain the parse tree. \\
    radtext-tree2dep & Parses to obtain the universal dependency graph. \\
    radtext-neg & Detects negations.\\
    radtext-collect & Collects and merges labels.\\
    radtext-csv2bioc & Converts CSV format to BioC format. \\
    radtext-cdm2bioc & Converts OMOP CDM format to BioC format. \\
    radtext-bioc2cdm & Converts BioC format to OMOP CDM format.\\
    \bottomrule
    \end{tabular}
    \vspace{.5cm}
    \caption{Command line commands.}
    \label{tab:radtext-commands}
\end{table}

\subsection{Python API Usage}
\radtext can be directly imported as a Python library. Users can access all the functionalities of \radtext through Python API.

\subsubsection{BioC-CDM Conversion}
\label{api-usage-convert}

\radtext's Python API supports the mutual conversion between BioC format and OMOP CDM. The following code snippet shows an example of converting BioC format to CDM and then converting CDM back to BioC format.

\begin{testexample}[An example of API usage]
\lstset{language=Python}
\begin{lstlisting}
import bioc
from radtext import BioC2CDM, CDM2BioC

# initialize RadText's BioC2CDM converter.
bioc2cdm = BioC2CDM()
with open(filepath) as fp:
    collection = bioc.load(fp)

cdm_df = bioc2cdm(collection)

# initialize RadText's CDM2BioC converter.
cdm2bioc = CDM2BioC()
bioc_collection = cdm2bioc(cdm_df)
\end{lstlisting}
\end{testexample}

\subsubsection{Pipeline Usage}

The following code snippet shows a minimal usage of \radtext's entire pipeline through Python API, which annotates a sample report and prints out all annotation results.

\begin{testexample}[An example of API usage]
\lstset{language=Python}
\begin{lstlisting}
import bioc
import radtext

# initialize RadText's pipeline.
nlp = radtext.Pipeline()

# load a BioC-format sample report.
with open(filepath) as fp:
    doc = bioc.load(fp)
    
# run RadText's pipeline on the sample report.
collection = nlp(doc)

print(collection)
\end{lstlisting}
\end{testexample}

After running all modules, \radtext returns a \texttt{Collection} instance that stores the final annotation results. Within a \texttt{Collection} instance, the annotations are stored in either \texttt{Passage} or \texttt{Sentence} classes. The following code snippet shows how we can access the detected disease findings and the corresponding negation status after obtaining the \texttt{Collection} instance.

\begin{testexample}[An example of API usage]
\lstset{language=Python}
\begin{lstlisting}
for doc in collection.documents:
    for passage in doc.passages:
        for annotation in passage.annotations:
            print(annotation.infon['source_concept'],
                    annotation.infon['negation'])
\end{lstlisting}
\end{testexample}

\radtext's Python API also allows partial pipeline execution. Therefore, users can pause after any module of \radtext to access the intermediate NLP results. The following code snippet shows an example of the partial execution of \radtext. By specifying the \texttt{annotators} to be \emph{secsplit} and \emph{ssplit}, \radtext will run section segmentation and sentence split sequentially. The output \texttt{Collection} instance will have the annotation results of sentence split. 

\begin{testexample}[An example of API usage]
\lstset{language=Python}
\begin{lstlisting}
import radtext

# initialize RadText's pipeline which will perform section segmentation and sentence split.
nlp = radtext.Pipeline(annotators=['secsplit', 'ssplit'])

# load a BioC-format sample report.
with open(filepath) as fp:
    doc = bioc.load(fp)
    
# run RadText's pipeline on the sample report.
collection = nlp(doc)

print(collection)
\end{lstlisting}
\end{testexample}

\section{Evaluation}
\subsection{Dataset}
We evaluated \radtext on the MIMIC-CXR dataset \cite{johnson2019mimic}. MIMIC-CXR is a large publicly available dataset of radiographic studies performed at the Beth Israel Deaconess Medical Center. This dataset contains 227,827 radiology reports in total.

\subsection{Experiments and Results}
We evaluated \radtext's performance on five new disease findings that were not covered by previous works, including Calcification of the Aorta, Pneumomediastinum, Pneumoperitoneum, Subcutaneous Emphysema, Tortuous Aorta.

\begin{table}[!htpb]
    \centering
    \small
    \begin{tabular}[width=0.5\textwidth]{lccc}
    \toprule
    Disease Finding & Precision & Recall & F-1 \\
    \midrule
    Calcification of the Aorta & 1.00 & 0.87 & 0.93 \\
    Pneumomediastinum & 0.70 & 1.00 & 0.82 \\
    Pneumoperitoneum & 0.88 & 1.00 & 0.94 \\
    Subcutaneous Emphysema & 0.95 & 0.91 & 0.93 \\
    Tortuous Aorta & 1.00 & 0.94 & 0.97 \\
    \midrule
    Macro Average & 0.91 & 0.94 & 0.92\\
    \bottomrule
    \end{tabular}
    \vspace{0.5cm}
    \caption{\radtext performances on five new disease findings.}
    \label{tab:f-1scores}
\end{table}

We randomly selected 200 test reports from the MIMIC-CXR dataset and manually annotated the five new disease findings. We evaluated \radtext by comparing the results of \radtext with the manually-annotated gold standard. Precision, recall, and F1-score were computed accordingly based on the number of true positives, false positives, and false negatives (see Table \ref{tab:f-1scores}). The average precision score is 0.91, with the highest precision being 1.0 for Calcification of the Aorta and Tortuous Aorta; the average recall score is 0.94, with the highest recall being 1.0 for Pneumomediastinum and Pneumoperitoneum; and the average F-1 score is 0.92, with the highest F-1 score being 0.97 for Tortuous Aorta. \radtext achieves an average precision of 0.91, an average recall of 0.94, and an average F-1 score of 0.92. All reports in the MIMIC-CXR dataset were analyzed using \radtext (see Table \ref{tab:mimic-radtext-results}). Among the five new disease findings, Calcification of the Aorta is mentioned in 3,380 reports, which Pneumoperitoneum is mentioned in only 1,604 reports. The labels can also be found at the \radtext homepage.

\begin{table}[ht!]
    \begin{tabular}{lrrrr}
    \toprule
    Finding & Positive & Negative & Uncertain & \textbf{Total} \\
    \midrule
    Calcification of the Aorta & 3,344 & 13 & 23 & 3,380 \\
    Pneumomediastinum & 779 & 856 & 131 & 1,766 \\
    Pneumoperitoneum & 580 & 938 & 86 & 1,604 \\
    Subcutaneous Emphysema & 2,529 & 131 & 31 & 2,691 \\
    Tortuous Aorta & 2,681 & 41 & 131 & 2,853 \\
    \bottomrule
    \end{tabular}
    \vspace{.5cm}
    \caption{Statistics of five new disease findings in MIMIC-CXR dataset.}
    \label{tab:mimic-radtext-results}
\end{table}

\section{Conclusion and Future Work}
In this work, we presented \radtext, a high-performance Python radiology text analysis system. We highlighted that \radtext features hybrid neural analysis, raw text processing and local processing, bringing better usability and data privacy. \radtext's modular design, user-friendly user interface, easy-to-use command line usage and Python APIs allow users to have great flexibility on the radiology text analysis task. We evaluated \radtext on the MIMIC-CXR dataset, especially on five new disease findings that were not covered by previous work, and the results demonstrated \radtext's superior performances on radiology report analysis.  \radtext employs BioC-CDM, which stores the results in the extended BioC format that is compatible with OMOP CDM. \radtext' compatibility with OMOP CDM supports collaborative research across disparate data sources. 

In the future, \radtext is going to be continuously maintained and expanded as new resources become available. For example, the NER module can be improved by incorporating scispaCy, developed for processing biomedical, scientific or clinical text \cite{neumann-etal-2019-scispacy}. By making \radtext publicly available, we envision it can facilitate future research and applications in the healthcare informatics community. 


\section*{Acknowledgment}

This work is supported by the National Library of Medicine under Award No. 4R00LM013001 and the NIH Intramural Research Program, National Library of Medicine. 

\bibliographystyle{IEEEtran}
\bibliography{conference_101719}

\begin{thebibliography}{10}
\providecommand{\url}[1]{#1}
\csname url@samestyle\endcsname
\providecommand{\newblock}{\relax}
\providecommand{\bibinfo}[2]{#2}
\providecommand{\BIBentrySTDinterwordspacing}{\spaceskip=0pt\relax}
\providecommand{\BIBentryALTinterwordstretchfactor}{4}
\providecommand{\BIBentryALTinterwordspacing}{\spaceskip=\fontdimen2\font plus
\BIBentryALTinterwordstretchfactor\fontdimen3\font minus
  \fontdimen4\font\relax}
\providecommand{\BIBforeignlanguage}[2]{{%
\expandafter\ifx\csname l@#1\endcsname\relax
\typeout{** WARNING: IEEEtran.bst: No hyphenation pattern has been}%
\typeout{** loaded for the language `#1'. Using the pattern for}%
\typeout{** the default language instead.}%
\else
\language=\csname l@#1\endcsname
\fi
#2}}
\providecommand{\BIBdecl}{\relax}
\BIBdecl

\bibitem{brady2016radiology-error}
A.~Brady, ``Error and discrepancy in radiology: inevitable or avoidable?''
  \emph{Insights into Imaging}, vol.~8, 12 2016.

\bibitem{savova2010ctakes}
G.~Savova, J.~Masanz, P.~Ogren, J.~Zheng, S.~Sohn, K.~Kipper-Schuler, and
  C.~Chute, ``Mayo clinical text analysis and knowledge extraction system
  ({cTAKES}): Architecture, component evaluation and applications,''
  \emph{JAMIA}, vol.~17, pp. 507--13, 09 2010.

\bibitem{neumann-etal-2019-scispacy}
M.~Neumann, D.~King, I.~Beltagy, and W.~Ammar, ``{S}cispa{C}y: {F}ast and
  {R}obust {M}odels for {B}iomedical {N}atural {L}anguage {P}rocessing,'' in
  \emph{Proceedings of the BioNLP Workshop and Shared Task}, Aug. 2019, pp.
  319--327.

\bibitem{liu2013medtagger}
H.~Liu, S.~J. Bielinski, S.~Sohn, S.~Murphy, K.~B. Wagholikar, S.~R.
  Jonnalagadda, K.~Ravikumar, S.~T. Wu, I.~J. Kullo, and C.~G. Chute, ``An
  information extraction framework for cohort identification using electronic
  health records,'' \emph{AMIA Joint Summits on Translational Science}, vol.
  2013, p. 149—153, 2013.

\bibitem{soysal2017clamp}
E.~Soysal, J.~Wang, M.~Jiang, Y.~Wu, S.~Pakhomov, H.~Liu, and H.~Xu, ``{CLAMP
  – a toolkit for efficiently building customized clinical natural language
  processing pipelines},'' \emph{JAMIA}, vol.~25, no.~3, pp. 331--336, 11 2017.

\bibitem{bodenreider2004unified}
O.~Bodenreider, ``The {{Unified Medical Language System}} ({{UMLS}}):
  Integrating biomedical terminology,'' \emph{Nucleic Acids Research}, vol.~32,
  no. Database issue, pp. D267--270, Jan. 2004.

\bibitem{langlotz2006radlex}
C.~P. Langlotz, ``{{RadLex}}: A new method for indexing online educational
  materials,'' \emph{Radiographics: A Review Publication of the Radiological
  Society of North America, Inc}, vol.~26, no.~6, pp. 1595--1597, 2006 Nov-Dec.

\bibitem{chapman2013extending}
W.~W. Chapman, D.~Hillert, S.~Velupillai, M.~Kvist, M.~Skeppstedt, B.~E.
  Chapman, M.~Conway, M.~Tharp, D.~L. Mowery, and L.~Deleger, ``Extending the
  {{NegEx}} lexicon for multiple languages.'' \emph{Studies in health
  technology and informatics}, vol. 192, pp. 677--681, 2013.

\bibitem{chapman2011documentlevel}
B.~E. Chapman, S.~Lee, H.~P. Kang, and W.~W. Chapman, ``Document-level
  classification of {{CT}} pulmonary angiography reports based on an extension
  of the {{ConText}} algorithm,'' \emph{Journal of Biomedical Informatics},
  vol.~44, no.~5, pp. 728--737, Oct. 2011.

\bibitem{fan2013syntactic}
J.-w. Fan, E.~W. Yang, M.~Jiang, R.~Prasad, R.~M. Loomis, D.~S. Zisook, J.~C.
  Denny, H.~Xu, and Y.~Huang, ``Syntactic parsing of clinical text: Guideline
  and corpus development with handling ill-formed sentences,'' \emph{JAMIA},
  vol.~20, no.~6, pp. 1168--1177, 2013 Nov-Dec.

\bibitem{peng2017negbio}
Y.~Peng, X.~Wang, L.~Lu, M.~Bagheri, R.~Summers, and Z.~Lu, ``Negbio: a
  high-performance tool for negation and uncertainty detection in radiology
  reports,'' in \emph{AMIA Joint Summits on Translational Science}, 2017, pp.
  188--196.

\bibitem{voss2015CDM}
E.~Voss, R.~Makadia, A.~Matcho, Q.~Ma, C.~Knoll, M.~Schuemie, F.~Defalco,
  A.~Londhe, V.~Zhu, and P.~Ryan, ``Feasibility and utility of applications of
  the common data model to multiple, disparate observational health
  databases,'' \emph{JAMIA}, vol.~22, 02 2015.

\bibitem{comeau2013bioc}
D.~Comeau, R.~Dogan, P.~Ciccarese, K.~Cohen, M.~Krallinger, F.~Leitner, Z.~lu,
  Y.~Peng, F.~Rinaldi, M.~Torii, A.~Valencia, K.~Verspoor, T.~Wiegers, C.~Wu,
  and W.~Wilbur, ``{BioC}: a minimalist approach to interoperability for
  biomedical text processing,'' \emph{Database : the journal of biological
  databases and curation}, vol. 2013, p. bat064, 01 2013.

\bibitem{pons2016naturala}
E.~Pons, L.~M.~M. Braun, M.~G.~M. Hunink, and J.~A. Kors, ``Natural {{Language
  Processing}} in {{Radiology}}: {{A Systematic Review}},'' \emph{Radiology},
  vol. 279, no.~2, pp. 329--343, May 2016.

\bibitem{aronson2010metamap}
A.~Aronson and F.-M. Lang, ``An overview of {MetaMap}: Historical perspective
  and recent advances,'' \emph{JAMIA}, vol.~17, pp. 229--36, 05 2010.

\bibitem{bodenreider2004umls}
O.~Bodenreider, ``The unified medical language system (umls): Integrating
  biomedical terminology,'' \emph{Nucleic acids research}, vol.~32, pp.
  D267--70, 02 2004.

\bibitem{eyre2021medspacy}
H.~Eyre, A.~B. Chapman, K.~S. Peterson, J.~Shi, P.~R. Alba, M.~M. Jones, T.~L.
  Box, S.~L. DuVall, and O.~V. Patterson, ``Launching into clinical space with
  medspacy: a new clinical text processing toolkit in python,'' in \emph{AMIA
  Annual Symposium Proceedings}, 2021.

\bibitem{honnibal2020spacy}
M.~Honnibal, I.~Montani, S.~Van~Landeghem, and A.~Boyd, ``spacy:
  Industrial-strength natural language processing in python,'' 2020.

\bibitem{Norgeot2020ProtectedHI}
B.~Norgeot, K.~Muenzen, T.~A. Peterson, X.~Fan, B.~S. Glicksberg, G.~Schenk,
  E.~Rutenberg, B.~Oskotsky, M.~Sirota, J.~Yazdany, G.~Schmajuk, D.~Ludwig,
  T.~Goldstein, and A.~J. Butte, ``Protected health information filter
  (philter): accurately and securely de-identifying free-text clinical notes,''
  \emph{NPJ Digital Medicine}, vol.~3, 2020.

\bibitem{rightsocr2012guidance}
O.~f.~C. Rights~(OCR), ``Guidance {{Regarding Methods}} for
  {{De}}-identification of {{Protected Health Information}} in {{Accordance}}
  with the {{Health Insurance Portability}} and {{Accountability Act}}
  ({{HIPAA}}) {{Privacy Rule}},''
  https://www.hhs.gov/hipaa/for-professionals/privacy/special-topics/de-identification/index.html,
  Sep. 2012.

\bibitem{tepper2012statistical}
M.~Tepper, D.~Capurro, F.~Xia, L.~Vanderwende, and M.~{Yetisgen-Yildiz},
  ``Statistical {{Section Segmentation}} in {{Free-Text Clinical Records}},''
  in \emph{Proceedings of the {{Eighth International Conference}} on {{Language
  Resources}} and {{Evaluation}} ({{LREC}})}, May 2012, pp. 2001--2008.

\bibitem{Wang2017ChestXRay8HC}
X.~Wang, Y.~Peng, L.~Lu, Z.~Lu, M.~Bagheri, and R.~M. Summers, ``Chestx-ray8:
  Hospital-scale chest x-ray database and benchmarks on weakly-supervised
  classification and localization of common thorax diseases,'' \emph{2017 IEEE
  Conference on Computer Vision and Pattern Recognition (CVPR)}, pp.
  3462--3471, 2017.

\bibitem{johnson2019mimic}
A.~Johnson, T.~Pollard, S.~Berkowitz, N.~Greenbaum, M.~Lungren, C.-y. Deng,
  R.~Mark, and S.~Horng, ``{MIMIC-CXR}, a de-identified publicly available
  database of chest radiographs with free-text reports,'' \emph{Scientific
  Data}, vol.~6, p. 317, 12 2019.

\bibitem{denny2008sectag}
J.~C. Denny, R.~A. Miller, K.~B. Johnson, and A.~Spickard, ``Development and
  evaluation of a clinical note section header terminology,'' in \emph{AMIA
  Symposium}, 2008, pp. 156--160.

\bibitem{mcdonald2003loinc}
C.~J. McDonald, S.~M. Huff, J.~G. Suico, G.~Hill, D.~Leavelle, R.~Aller,
  A.~Forrey, K.~Mercer, G.~DeMoor, J.~Hook, W.~Williams, J.~Case, and
  P.~Maloney, ``{{LOINC}}, a universal standard for identifying laboratory
  observations: A 5-year update,'' \emph{Clinical Chemistry}, vol.~49, no.~4,
  pp. 624--633, Apr. 2003.

\bibitem{miller1989use}
R.~A. Miller and F.~E. Masarie, ``Use of the {{Quick Medical Reference}}
  ({{QMR}}) program as a tool for medical education,'' \emph{Methods of
  Information in Medicine}, vol.~28, no.~4, pp. 340--345, Nov. 1989.

\bibitem{bird2009NLTK}
S.~Bird, E.~Klein, and E.~Loper, \emph{Natural Language Processing with
  Python}, 1st~ed.\hskip 1em plus 0.5em minus 0.4em\relax O'Reilly Media, Inc.,
  2009.

\bibitem{qi2020stanza}
P.~Qi, Y.~Zhang, Y.~Zhang, J.~Bolton, and C.~D. Manning, ``Stanza: A {Python}
  natural language processing toolkit for many human languages,'' in
  \emph{Proceedings of ACL: System Demonstrations}, 2020.

\bibitem{honnibal-johnson-2015-improved}
M.~Honnibal and M.~Johnson, ``An improved non-monotonic transition system for
  dependency parsing,'' in \emph{Proceedings of EMNLP}, Sep. 2015, pp.
  1373--1378.

\bibitem{nadeau2007ner}
D.~Nadeau and S.~Sekine, ``A survey of named entity recognition and
  classification,'' \emph{Lingvisticae Investigationes}, vol.~30, 08 2007.

\bibitem{charniak-johnson-2005-coarse}
E.~Charniak and M.~Johnson, ``Coarse-to-fine n-best parsing and {M}ax{E}nt
  discriminative reranking,'' in \emph{Proceedings of ACL}, Jun. 2005, pp.
  173--180.

\bibitem{Charniak2010AnyDP}
E.~Charniak and D.~McClosky, ``Any domain parsing: automatic domain adaptation
  for natural language parsing,'' 2010.

\bibitem{de-marneffe-etal-2014-universal}
M.-C. de~Marneffe, T.~Dozat, N.~Silveira, K.~Haverinen, F.~Ginter, J.~Nivre,
  and C.~D. Manning, ``Universal {S}tanford dependencies: A cross-linguistic
  typology,'' in \emph{Proceedings of the Ninth International Conference on
  Language Resources and Evaluation ({LREC})}, May 2014, pp. 4585--4592.

\bibitem{marneffe2008dependency}
M.-C. Marneffe and C.~Manning, ``The {Stanford} typed dependencies
  representation,'' \emph{COLING Workshop on Cross-framework and Cross-domain
  Parser Evaluation}, 01 2008.

\bibitem{Chapman2001EvaluationON}
W.~W. Chapman, W.~Bridewell, P.~Hanbury, G.~F. Cooper, and B.~G. Buchanan,
  ``Evaluation of negation phrases in narrative clinical reports,''
  \emph{Proceedings. AMIA Symposium}, pp. 105--9, 2001.

\end{thebibliography}

\end{document}